\documentclass[letterpaper, 10 pt, conference]{ieeeconf}
\usepackage[pdftex]{graphicx}

\usepackage{xcolor}
\usepackage{cite}

\usepackage{amsmath}
\usepackage{amssymb}
  \usepackage[caption=false,font=normalsize,labelfont=sf,textfont=sf]{subfig}
\usepackage{url}
\begin{document}
\title{Physics Models for Sim-to-Real Transfer\\in Professional-Level Robot Table Tennis}

\author{Christian Conti$^{1}$,
        Bilan Yang$^{1}$,
        Alexander Sigrist$^{2}$,
        Lorenzo Miele$^{2}$,\\
        Yamen Saraiji$^{1}$,
        Peter D\"urr$^{2}$,
        Naoya Takahashi$^{2}$ % <-this % stops a space
\\
\\
$^{1}$Sony AI, Tokyo, Japan \\
$^{2}$Sony AI, Z\"urich, Switzerland
}

\maketitle

\begin{abstract}
At competitive speeds and spins, a table tennis ball follows complex, counterintuitive trajectories that a robot must track and precisely counter within fractions of a second.
Training a reinforcement learning policy capable of these skills is prohibitively expensive and dangerous in the real world, making high-fidelity simulation essential. Transferability of such policies, however, critically depends on how faithfully the simulation captures real-world dynamics—a requirement made even more stringent by the adversarial nature of the game, where any regime in which a model fails to approximate reality becomes an exploitable weakness for the opponent.
Prior state-of-the-art in robot table tennis generally focuses on a limited range of velocities and spins and fails to capture the richness of ball behaviors encountered in professional-level play.
In this work, we present physics models for the aerodynamic ball flight, for the contact dynamics between the ball and the table, as well as between the ball and the racket that accurately capture the ball behavior over a vast range of speeds and spins relevant to the game. Specifically, we model drag and Magnus force coefficients as functions of Reynolds number and spin ratio in the aerodynamics equations. For the table contact model we model effects of ball buckling on the coefficient of restitution and incorporate residuals into the instantaneous point-contact models. For the racket contact model, we introduce a residual neural network component to complement coefficients related to normal and tangential coefficients of restitution as well as torsional spin damping.
Evaluated on an unprecedentedly large dataset of competitive matches (277 games), the proposed models significantly reduces the prediction errors (e.g., 59\%  median landing-position error reduction). 
The resulting models were used for the first real-world robot table tennis AI agent capable of competing against professional players, to train reinforcement learning policies.\footnote{\url{https://ace.ai.sony/}}
%The resulting models were used to train reinforcement learning policies on a custom-built robot that has been shown to defeat some of the world's top players, as initially demonstrated in \cite{Durr2026} and follow-up work\footnote{\url{https://ace.ai.sony/}}.

%Due to the prohibitive costs of training RL policies in real world, physics models are required to simulate the system in a virtual environment. A policy trained in such an environment will inevitably suffer from transferability issues due to the gap between simulated environments and reality. To reduce this gap, fast and accurate physics modeling of table tennis ball dynamics is necessary. In this work we present the models that have been developed for the robotic table tennis system presented in \cite{Nature}.
\end{abstract}

%\begin{IEEEkeywords}
%Robot table tennis, sim-to-real gap, residual neural network
%\end{IEEEkeywords}

\IEEEpeerreviewmaketitle

\section{Introduction}
% the challenge of competitive table tennis
Competitive table tennis demands fast perception and reaction: a player has a fraction of a second to interpret the trajectory of a ball traveling up to 35 m/s \cite{Tang2002} and return it precisely. Spin, which can reach 1000 rad/s \cite{Tang2002}, adds further complexity by producing counterintuitive trajectories through the Magnus force and contact physics. A heavy backspin ball, for example, falls more slowly than a low-spin ball and can bounce backward upon table contact, while imparting a strong downward force on the racket that the player must compensate for.
Reinforcement learning (RL) is a natural fit for controlling a robot at this level, but training in the real world is prohibitively expensive and unsafe. Training is therefore conducted in simulation, making the fidelity of the physics model crucial for sim-to real transfer. The sim-to-real gap is further exacerbated by the adversarial nature of the game: a skilled opponent will actively search for trajectories the robot cannot handle, turning minor modeling inaccuracies into systematic, exploitable weaknesses.
Several physics models are proposed for table tennis ball dynamics, including model based approaches \cite{Nakashima2011}, and data-driven  approaches \cite{Zhao2016,Zhao2017,Lin2020,Baptiste2024,Yang2021}. 
However, they focus on a limited range of velocities and spins and do not cover the full range observed in elite play.
% Model-based aerodynamics rely on constant coefficients that break down outside a narrow regime, and contact models miss effects that emerge at high impact speeds and spins. Data-driven approaches, in turn, require large datasets covering the deployment distribution and generalize poorly to rare or adversarial trajectories. 
%contribution of this work
In this work, we present improved physics models for aerodynamic flight, table contact, and racket contact that remain accurate across this full range while being computationally efficient for RL training.  The models are developed and validated iteratively in actual competitive games of elite and professional players, with large data collected at 200 Hz using nine cameras and three gaze control systems (GCSs) \cite{Hu2026} for accurate ball state measurement.
The aerodynamics model is improved with drag and Magnus coefficients that are modeled as functions of Reynolds number and spin ratio.
The contact models are improved by considering surface contact effects through improved restitution coefficients and residual models, notably using a neural-network in the racket contact model to handle complicated racket properties and inhomogeneities.
We evaluate the proposed models against state-of-the-art baselines on the large robot table tennis game data against elite and professional players and show that each reduces prediction error, narrowing the sim-to-real gap that skilled opponents could otherwise exploit in professional-level play.
The proposed models reduce the median landing-position error by 59\% over the standard baseline \cite{Nakashima2011}, significantly closing a sim-to-real gap.
% In this work, we present a set of physics models covering the aerodynamic flight of the ball, table contact, and racket contact, developed and validated iteratively in actual competitive games, with data collected at 200 Hz using nine cameras and three gaze control systems (GCSs) \cite{Hu2026} for accurate measurements.
% from the same perception system used in deployment.
% The models are designed to accurately capture the range of trajectories that can appear at highly competitive levels of play while remaining computationally efficient such that they can be used in an RL training environment.
% The aerodynamics model is therefore improved with drag and Magnus coefficients that are modeled as functions of Reynolds number and spin ratio, and the contact models include surface contact effects through improved restitution coefficients and residual models.
% We achieve these results by modeling parameters based on the data we collected (and qualitatively validated through literature) and by using a neural-network residual model in the racket contact model to handle complicated racket properties and inhomogeneities.
% We compare the proposed models against state-of-the-art approaches and show that each reduces prediction error over standard baselines — closing sim-to-real gaps that, at professional level, would otherwise be exploited by skilled opponents.
These models represent an advancement over D\"urr et al. \cite{Durr2026} and were used to train RL policies deployed in matches against progressively stronger opponents, including some of the best players in the world.

The main contributions of this work are three physics models developed and evaluated on an unprecedentedly large dataset of elite and professional game play data (277 games, 58,000 flight segments, 25,000 table contacts, and 11,000 racket contacts), enabling accurate characterization across the wide range of velocities and spins encountered in competitive table tennis: 
\begin{itemize}
    \item An aerodynamics model in which drag and Magnus coefficients are functions of Reynolds number and spin ratio, enabling accurate modeling across the wide range of velocities and spins encountered in competitive table tennis (up to $35$~m/s and $1000$~rad/s respectively \cite{Tang2002}).
    \item A table contact model that accounts for the effect of ball buckling on the coefficient of restitution and incorporates a residual correction for systematic deviations from the idealized point-contact model.
    \item A racket contact model complemented with a residual neural network that captures nonlinear behavior of racket-ball interaction.
\end{itemize}

\section{Related work}
Robot table tennis is a benchmark for fast perception and control and since 1983 \cite{billingsley1983robot} there has been a growing body of work tackling this challenge. Consequently, a vast amount of work is also available on the modeling of the ball physics.
In an effort to simplify the problem, in many cases spin has been largely ignored \cite{Huang2015,Bchler2020,Nguyen2025,Su2025,Ma2023} or restricted to a lower amount by hampering the player \cite{DAmbrosio2024}, even though it is a fundamental characteristic of competitive table tennis \cite{Tebbe2019}.
Any full model of the ball physics must address three tightly coupled main components: aerodynamic flight, table contact, and racket contact. 
Nakashima et al. \cite{Nakashima2011} established the most commonly used set of models, which use a standard aerodynamics model with gravity, aerodynamic drag, and the Magnus force — the lift force acting on a spinning ball moving through air — and complement them with contact models derived from simplified assumptions of instantaneous point contacts. Their proposed table contact model further distinguishes between rolling and sliding regimes and their racket contact model uses a linear elastic interaction with a tangential impulse proportional to the pre-impact surface velocity. Although these models work sufficiently well in simpler settings, the range of velocities and spins observed in professional table tennis is considerably larger and assumptions of constant drag and Magnus coefficients, as well as instantaneous point contacts are pushed beyond their limits. The adversarial nature of the game further contributes to invalidating these models.
%These models therefore form the baseline against which the present work is compared.

D\"urr et al. \cite{Durr2026} extended this framework to handle the wide range of velocities and spins encountered in professional-level play, introducing a velocity-dependent Magnus coefficient, a velocity-dependent table restitution coefficient, and a small residual neural network correction for the racket contact model. The present work builds directly on these models, advancing them further with richer aerodynamic coefficient representations and an improved racket contact formulation; both Nakashima et al. \cite{Nakashima2011} and D\"urr et al. \cite{Durr2026} are therefore used as baselines throughout our evaluation.

Beyond white-box physics models, several approaches incorporate data-driven components at varying levels of abstraction.
At one extreme, purely black-box methods approximate ball trajectories with polynomial curves or neural networks fitted to recorded data, avoiding explicit assumptions about the physics~\cite{Zhao2016,Zhao2017,Lin2020,Bi2026}.
%While computationally lightweight, 
However, these approaches require large datasets covering the deployment distribution and do not generalize reliably to rare or adversarially constructed trajectories.
Intermediate gray-box approaches combine a physics backbone with learned corrections or parameters.
Notably, Baptiste and Maxime \cite{Baptiste2024} combine a physics-based flight model with a machine-learning component for trajectory prediction in table tennis, and Yang et al. \cite{Yang2021} employ deep reinforcement learning for stroke control with LSTM-based spin estimation built on top of Nakashima et al.'s contact models.
More broadly, hybrid frameworks that augment analytical models with residual networks or sparse regression have shown promise for scientific and robotic applications~\cite{Kaheman2019,Liang2020}, and physics-informed neural networks offer a complementary direction for incorporating physical constraints into learned models~\cite{Raissi2017}.
The present work follows the gray-box philosophy: we retain interpretable white-box models as the backbone and extend them with data-fitted functional forms and residual neural networks, with the specific aim of reducing the sim-to-real gap in an adversarial competitive setting.

Other studies address single components in more detail, e.g. via finite element simulations for the table contact \cite{Bao2015}, computational fluid dynamics simulations \cite{Ito2025}, and wind tunnel experiments \cite{Miyazaki2017}. These approaches are however extremely computationally expensive and therefore not practical for the full game simulations required to train RL policies.
Other relevant publications with real world experiments look at the oblique impact of a ball with a flat surface \cite{Rmond2023}, and at the change in coefficient of restitution of the racket at different impact speeds due to buckling \cite{Cross2014}.

% In this work we revisit each of the three models presented in \cite{Nakashima2011} by extending them based on data we collected during the project development, to work in a competitive environment where a large sim-to-real gap becomes an exploitable weakness for the robot.
% In the aerodynamics model, the definition of drag and Magnus coefficients are updated to fit the large range of velocities (up to $30$ m/s) and spins (up to $1000$ rad/s) that we encounter in actual matches by moving from constant parameters to fitted functions of Reynolds number $Re$ and spin ratio $Sp$.
% The table contact model has been slightly updated by moving from a constant coefficient of restitution to a linear restitution model and by adding residual correction matrices.
% Finally, the racket contact model has undergone more important updates to handle non-constant coefficients for the normal coefficient of restitution, the tangential coefficient of restitution and the additional torsional damping on normal spin. The model is further complemented by a neural network-based residual model to capture more complex, unmodeled interactions (such as observed variations in physical properties across the racket).

% The rest of this article is structured as follows: Section \ref{Method} will discuss the details of the three proposed physics models, then Section \ref{Results} will show results to justify the models.  Finally Section \ref{conclusion} will conclude this article.

\section{Method}
\label{Method}
We consider the problem of constructing computationally efficient physics models for aerodynamic ball flight, table contact, and racket contact in competitive table tennis. Given the ball state (position, linear velocity, and angular velocity) and racket state (position, orientation, linear velocity, and angular velocity) the models predict its evolution during free flight and after each contact event. Once fit to real-world data, these models generate synthetic trajectories used to train reinforcement learning policies that transfer zero-shot to real competitive play. Each model is evaluated individually against  observations from real matches, and then assessed jointly through landing position error over full simulated rally segments.
%In this section, we first describe the data used in the models, followed by the aerodynamics flight model, table contact model, and finally the racket contact model.
% This Section will briefly discuss the data used in the models which will be subsequently presented starting with the aerodynamics flight model, followed by the table contact model, and finishing with the racket contact model.
% In this work, we do not model net contacts, as they are difficult to simulate accurately. Nevertheless, the high re-planning frequency of the RL policy enables rapid adaptation to updated observations, in practice allowing the system to recover from unexpected contact events.

\subsection{Data collection}
\label{section:uncertainties}
% Ball trajectories from multiple matches against players of all levels have been collected over multiple sessions of real play with the perception system described in \cite{Durr2026}.
Ball trajectory data is collected with the competitive table tennis robot described in D\"urr et al. \cite{Durr2026},  from multiple matches against a wide range of players including amateur, elite, and professional players.
This dataset contains $277$ games that have been segmented for various contact events, for a total of almost $58,000$ flight trajectories, almost $25,000$ table contacts and $11,000$ robot racket contacts.
All ball trajectory data is recorded using nine cameras for wide coverage and accurate triangulation and three gaze control systems \cite{Hu2026} for accurate spin measurement. Robot racket states are obtained from encoder readings via forward kinematics.  All equipment is approved and certified by the International Table Tennis Federation (ITTF): Nittaku Nexcel 40+ 3 star balls, a SAN-EI table, and rackets consisting of VICTAS ZX-GEAR OUT\texttrademark\ blades with Butterfly Dignics05 2.1mm rubbers.
% The data contains multiple uncertainties that affect the results:
% \begin{itemize}
%     \item equipment wear and tear (especially racket rubbers),
%     \item damaged balls (while they get regularly replaced, during single games slightly damaged balls can be used inadvertently),
%     \item table playing surface, which although meets ITTF standards, is still not perfectly flat and level, partially also due to wear and tear, %: we observe that a 0.5 deg deviation in table normal vector can produce errors of up XXX in landing position
%     \item spin measurement errors (while the estimation method is generally reliable, e.g. due to occlusions, it can sometime yield incorrect values),
%     \item robot vibration: while we compute the racket position from forward kinematics based on encoder readings, we observed that the robot can have a significant amount of vibrations, which will affect the racket in estimation of the local frame and also local contact quantities
%     \item air flow: while minimal, we observed that it does affect reproducibility of shots such as serves.
% \end{itemize}

% To reduce the effect of bad data on the modeling, data is filtered as follows:
% \begin{itemize}
%     \item trajectories where spin estimates are less reliable are excluded (flight segments with a duration of less than $0.15$ s or flight segments for which the confidence of the spin estimation is low)
%     \item trajectories for which the RMSE error is larger than $20$ mm to exclude segmentation errors, but generally not discrepancies between model and observations
% \end{itemize}

Despite conducting matches under ITTF regulations and highly accurate perception systems, the data can still contain multiple uncertainties due to occasional occlusion, segmentation label error, structural vibrations in the robot, equipment wear and tear, temporal system instability, and air flow.
To make sure clean data is used for modeling, we filtered the data by excluding flight segments (1) with a low confidence score in the spin estimation \cite{Hu2026} and (2) trajectories for which the RMSE error between the observations and the fitted aerodynamics model is greater than $20$ mm, which correspond to segmentation label errors, but generally not discrepancies between model and observations.

\subsection{Aerodynamics Model}
As done in previous work \cite{Nakashima2011},  due to the physics involved in ball flight, we consider an aerodynamics model that includes drag, Magnus and gravitational forces:
\begin{eqnarray}
m\mathbf{a} &=& \mathbf{F}_D + \mathbf{F}_M + \mathbf{F}_G \\
&=& -\frac{1}{2} C_D \rho_{A} A \mathbf{v} \| \mathbf{v} \| - C_M \rho_{A} V \mathbf{v} \times \boldsymbol{\omega} + m\mathbf{g} ,
\end{eqnarray}
where $A=r^2\pi$ is the projected area of the ball (of radius $r=0.02$m), $V=\frac{4}{3} r^3\pi$ is the ball volume,  $\rho_A$ is the air density, $\mathbf{g}$ is the gravity vector, $m$ is the mass of the ball and $\mathbf{a}$, $\mathbf{v}$, $\boldsymbol{\omega}$ are the ball acceleration, linear velocity and angular velocity (which we will use interchangeably with spin, as it is the common term used in table tennis) respectively. $C_D$ and $C_M$ denote the drag and Magnus coefficients respectively.

We improve on previous work by modeling $C_D$ and $C_M$ as functions of Reynolds number and spin ratio since constant coefficients are not sufficient to describe the complex motion of a flying and spinning table tennis ball. 
In fact, from fluid mechanics and experimental observations we know that these quantities generally depend on the Reynolds number $Re=\frac{2vr}{\nu}$ --- a dimensionless ratio of inertial to viscous forces in the flow --- and the spin ratio $Sp=\frac{r\omega}{v}$, both of which vary substantially in competitive table tennis. Here $\nu = 1.506 \times 10^{-5}$~m$^2$/s is the kinematic viscosity of air at $20^\circ$C.
We determine $C_D=C_D(Re,Sp)$ and $C_M=C_M(Re,Sp)$ by first fitting coefficients to each individual trajectory, then regressing these optimized coefficients against $Re$ and $Sp$. For each trajectory, the optimal coefficients are associated with an effective velocity $\mathbf{v}_{\mbox{\emph{eff}}}$, defined as the ``force-averaged" velocity over the segment - the constant velocity that would produce the same drag or Magnus impulse as the actual trajectory.
%We find this relation by finding the coefficients that fit the individual trajectories we collected and then by fitting the functions $C_D=C_D(Re,Sp)$ and $C_M=C_M(Re,Sp)$ to these optimized coefficients where, for each trajectory, the optimal set of coefficients is associated to an effective velocity $\mathbf{v}_{\mbox{\emph{eff}}}$ which is a force-averaged velocity over the segment and is the velocity that, if kept constant during flight, would yield the same drag impulse as the actual velocity.
For the drag force, 
\begin{equation}
    v_{\mathrm{eff}}=\frac{\sum_i \| \mathbf{v}\| _i^3}{\sum_i \| \mathbf{v}\| _i^2},
\end{equation}
and for the Magnus force,
\begin{equation}
    v_{\mathrm{eff}}=\frac{\sum_i \| \mathbf{v}\| _i^2 \sin\theta_i}{\sum_i ||\mathbf{v}\| _i \sin\theta_i}
\end{equation}
where $\sin\theta_i=\frac{\| \mathbf{v} \times \boldsymbol{\omega}\|_i }{\| \mathbf{v}\|_i  \| \boldsymbol{\omega}\|_i }$.
%As spin $\boldsymbol{\omega}$ is generally considered \cite{} (and observed) constant during flight, there is no need to calculate an effective spin.

\subsubsection{Drag coefficient $C_D$}
\begin{figure}
    \centering
    \includegraphics[width=\linewidth]{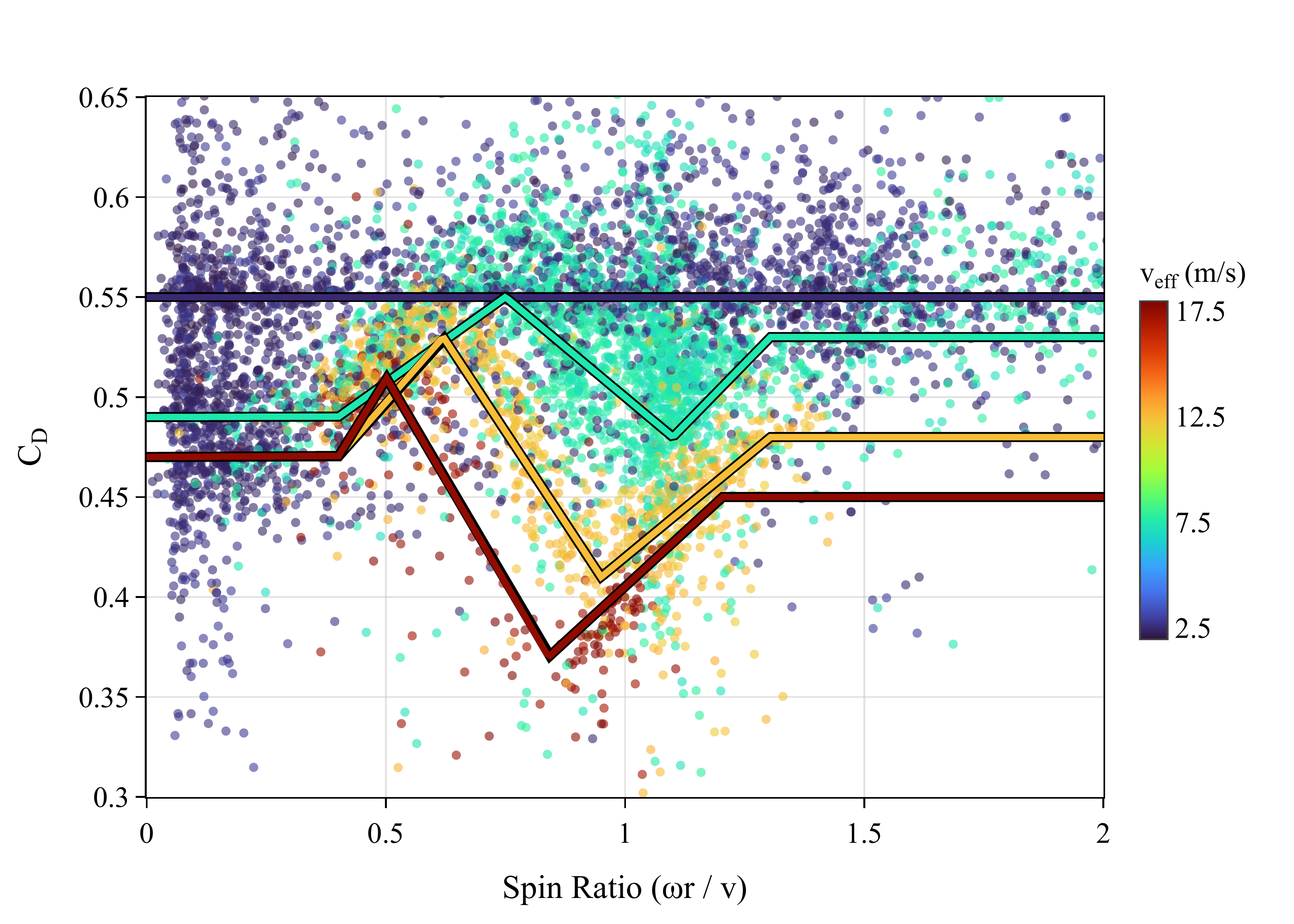}
    \caption{Drag coefficient vs.\ $Sp$ for $v_{\mathrm{eff}} \in \{2.5, 7.5, 12.5, 17.5\}~\mathrm{m/s}$: red hues indicate higher velocity, blue hues lower velocity.}
    \label{fig:drag}
\end{figure}

The drag coefficient model is based on the observation that for a fixed Reynolds number $Re^{(k)}$, the optimal fit for the drag coefficient with respect to the spin ratio $Sp$ can be approximated by a piecewise linear function with a ridge and a trough (see Fig. \ref{fig:drag}).
We therefore construct the model by choosing 4 values of $Re^{(k)} \in \{6{,}640;\; 19{,}920;\; 33{,}201;\; 46{,}481\}$ (corresponding to $v^{(k)} = Re^{(k)}\nu/2r \in \{2.5,\; 7.5,\; 12.5,\; 17.5\}$~m/s) over which the piecewise linear functions are built and then linearly interpolate between them to cover the whole range of $Re$ that we are interested in.
The piecewise linear function is defined at six breakpoints $\{Sp_{0}^{(k)}, \dots, Sp_{5}^{(k)}\}$ with corresponding values $\{C_{D,0}^{(k)}, \dots, C_{D,5}^{(k)}\}$ (see Table \ref{tab:drag_breakpoints} for the values).
In between these $v_{\mathrm{eff}}^{(k)}$ values, $C_D$ is linearly interpolated.
Furthermore, for $v_{\mathrm{eff}}^{(k)}$ larger than $17.5$ m/s and up to around $30$ m/s, $C_D$ is linearly extrapolated: given the short duration of high velocity trajectories (generally corresponding to smashes or fast drive shots), small errors in drag coefficient do not have time to integrate and cause large errors in trajectory predictions.

\begin{table}[t]
\caption{Drag coefficient breakpoints $C_D^{(k)}$ at spin ratio $Sp^{(k)}$ 
         for each reference velocity $v^{(k)}$ (m/s).}
\label{tab:drag_breakpoints}
\centering
\begin{tabular}{c l l l l l l}
 $v_{\mathrm{eff}}^{(k)}$ & {$Sp^{(k)}_0$} & {$Sp^{(k)}_1$} & {$Sp^{(k)}_2$} & {$Sp^{(k)}_3$} & {$Sp^{(k)}_4$} & {$Sp^{(k)}_5$} \\
  & {$C_{D,0}^{(k)}$} & {$C_{D,1}^{(k)}$} & {$C_{D,2}^{(k)}$} & {$C_{D,3}^{(k)}$} & {$C_{D,4}^{(k)}$} & {$C_{D,5}^{(k)}$} \\
\hline \hline
 $2.5$& $0$ & $0.3$ & $0.7$ & $0.95$ & $1.5$ & $2.0$\\
     & $0.55$ & $0.55$ & $0.55$ & $0.55$ & $0.55$ & $0.55$ \\
\hline
 $7.5$ & $0$ & $0.4$ & $0.75$ & $1.1$ & $1.3$ & $2.0$\\
     & $0.49$ & $0.49$ & $0.55$ & $0.48$ & $0.53$ & $0.53$ \\
\hline
 $12.5$ & $0$ & $0.4$ & $0.62$ & $0.95$ & $1.3$ & $2.0$\\
     & $0.47$ & $0.47$ & $0.53$ & $0.41$ & $0.48$ & $0.48$ \\
\hline
 $17.5$& $0$ & $0.4$ & $0.5$ & $0.84$ & $1.2$ & $2.0$\\
     & $0.47$ & $0.47$ & $0.51$ & $0.37$ & $0.45$ & $0.45$ \\
\end{tabular}
\end{table}
Fig. \ref{fig:drag} shows the fitted function with the corresponding values estimated from observed trajectories in a range of $\pm 0.5$ m/s. 
At low velocity the drag coefficient tends to $C_D=0.55$, a value similar to the one observed by Nakashima et al. \cite{Nakashima2011}.
The plot also shows how at lower velocities the estimates of the drag coefficients present a high variance due to the lower influence of the drag on the trajectories and the higher relative triangulation errors.
Note that the data is generally concentrated in the ridge and trough regions, at high velocity there is considerably less data and at low velocity the data is much noisier (but also less important to capture correctly as inaccurate estimates in this regime have negligible influence over the trajectory and thus are not exploitable).
%The aerodynamics coefficients proposed in this work are qualitatively comparable with the work of Miyazaki et al. \cite{Miyazaki2017}, which observes a similar behavior in experiments (where they consider $Re \in [30000;90000]$ and $Sp \in [0;1]$), and Ito et al. \cite{Ito2025} which shows an analysis done with Large Eddy Simulations over $Re \in [6600;53000]$ and $Sp \in [0.1;1.2]$ (and only a few data points at higher Sp for low $Re$). Both works however cover a smaller range of $Re$ and $Sp$.
The identified aerodynamic coefficients are qualitatively consistent with prior experimental and CFD studies \cite{Miyazaki2017}, \cite{Ito2025}, while covering a wider range of Reynolds numbers and spin ratios.

\subsubsection{Magnus coefficient $C_M$}
\begin{figure}
    \centering
    \includegraphics[width=\linewidth]{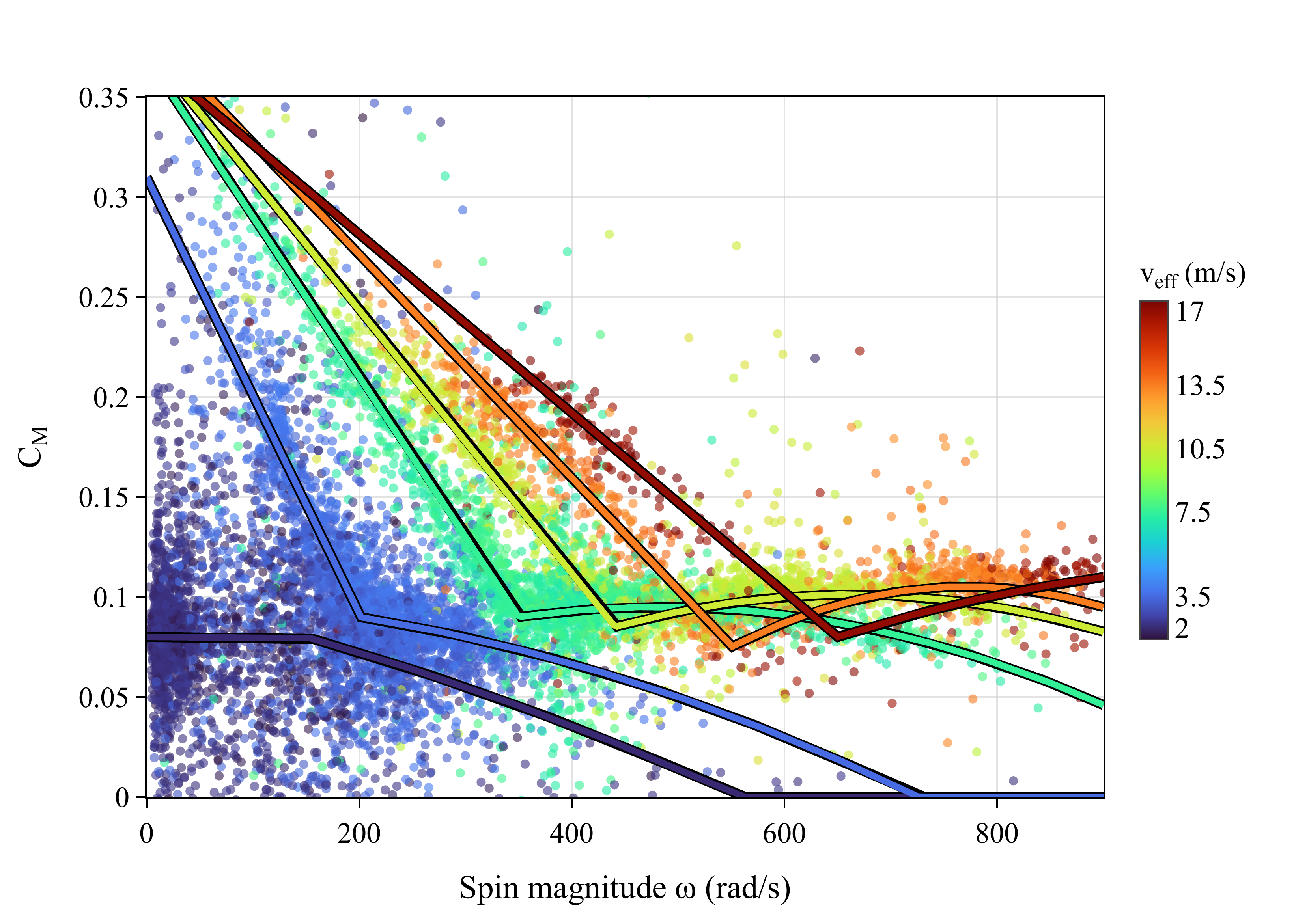}
    \caption{Magnus coefficient vs.\ $\omega$ for $v_{\mathrm{eff}} \in \{2.0, 3.5, 7.5, 10.5, 13.5, 17.0\}~\mathrm{m/s}$: red hues indicate higher velocity, blue hues lower velocity.}
    \label{fig:magnus}
\end{figure}

We observe from the data that at each fixed Reynolds number $Re^{(k)}$, the behavior of the Magnus coefficient can be represented by a piecewise function composed of a linearly decreasing segment and a negative quadratic part with the transition point from one to the other being $\omega_b$ (see Fig.\ref{fig:magnus}).

Therefore we express $C_M$ as a function of $v$ and $\omega$ (trivially convertible to $C_M(Re, Sp)$ via $v = Re\,\nu/2r$ and $\omega = Sp\,Re\,\nu/2r^2$):
\begin{equation}
  C_M(v, \omega) =
  \begin{cases}
    m_1\,\omega + s
      & \omega \leq \omega_b \\[6pt]
    a\,\omega^2 + b\,\omega + c
      & \omega > \omega_b
  \end{cases}
  , \qquad C_M \geq 0
\end{equation}

with the fitted parameters for the linear and quadratic components shown in Table \ref{tab:magnus_linear} and in Table \ref{tab:magnus_quadratic} respectively.

\begin{table}[t]
\caption{Coefficients $m_1$ and $s$ of the linear component of $C_M$.}
\label{tab:magnus_linear}
\centering
\begin{tabular}{c c c c }
\hline
$v_{\mathrm{eff}}^{(k)}$~[m/s]& $m_1$ & $s$& $\omega_b$~[rad/s]  \\
\hline
$2.0$  & $0$                    & $0.08$  & $150$ \\
$3.5$  & $-1.1 \times 10^{-3}$  & $0.31$  & $200$  \\
$7.5$  & $-8.0 \times 10^{-4}$  & $0.37$  & $350$  \\
$10.5$ & $-6.58 \times 10^{-4}$ & $0.375$ & $440$  \\
$13.5$ & $-5.6 \times 10^{-4}$  & $0.383$ & $550$  \\
$17.0$ & $-4.48 \times 10^{-4}$ & $0.371$ & $650$  \\
\hline
\end{tabular}
\end{table}

\begin{table}[t]
\caption{Coefficients $a$, $b$ and $c$ of the quadratic component of $C_M$.}
\label{tab:magnus_quadratic}
\centering
\begin{tabular}{c c c c}
\hline
$v_{\mathrm{eff}}^{(k)}$~[m/s]& a & b & c\\
\hline
$2.0$  & $-1.852 \times 10^{-7}$ & $-1.296 \times 10^{-4}$ & $0.0983$\\
$3.5$  & $-1.667 \times 10^{-7}$ & $-3.333 \times 10^{-5}$&$0.1$\\
$7.5$  & $-2.000 \times 10^{-7}$ & $1.700 \times 10^{-4}$&$0.0587$\\
$10.5$ & $-2.604 \times 10^{-7}$ & $3.646 \times 10^{-4}$&$-0.0225$\\
$13.5$ & $-3.571 \times 10^{-7}$ & $5.357 \times 10^{-4}$&$-0.0893$\\
$17.0$ & $-1.000 \times 10^{-7}$ & $2.300 \times 10^{-4}$&$-0.0375$\\
\hline
\end{tabular}
\end{table}

%Note that here we express $C_M$ as a function of $v$ and $\omega$ as it results in a simpler form. However, conversion to $C_M(Re, Sp)$ is trivial by substituting $v$ with $\frac{Re~\nu}{2r}$ and $\omega$ with $Sp~Re~\frac{\nu}{2r^2}$ and computing the corresponding $\omega_b$.
Outside of the $Re^{(k)}$ range defined, the function is clipped.

Fig. \ref{fig:magnus} shows the $C_M$ lines for the chosen $v_{\mathrm{eff}}^{(k)}$.
As with the drag coefficient, low-velocity Magnus coefficient estimates exhibit higher variance because triangulation errors have a larger impact on trajectory fitting.

%{\color{red}TODO update models? the lift plots show more space for improvement that is not as clear in Magnus plots. However currently it looks like we're roughly within $5$ cm in landing point. The update would be mostly for presentation purposes and I don't expect it to be very impactful}

\subsection{Table Contact Model}
Following up on Nakashima et al. \cite{Nakashima2011}, we start from the following table contact model, which models contacts as instantaneous point contacts:
\begin{equation}
  \mathbf{v}^+ = \mathbf{A}_{vv}\,\mathbf{v}^- + \mathbf{A}_{v\omega}\,\boldsymbol{\omega}^-, \qquad
  \boldsymbol{\omega}^+ = \mathbf{A}_{\omega v}\,\mathbf{v}^- + \mathbf{A}_{\omega\omega}\,\boldsymbol{\omega}^-
  \label{eq:table_nakashima}
\end{equation}
with
\begin{align}
\label{eq:tcm_v}
\resizebox{\columnwidth}{!}{$
\mathbf{A}_{vv} = \begin{pmatrix}
1-\alpha & 0 & 0\\
0 & 1-\alpha & 0\\
0 & 0 & -e_n
\end{pmatrix}
, \quad
\mathbf{A}_{v \omega} = \begin{pmatrix}
0 & \alpha r & 0\\
-\alpha r & 0 & 0\\
0 & 0 & 0
\end{pmatrix}
$},
\end{align}
\begin{align}
\label{eq:tcm_w}
\resizebox{\columnwidth}{!}{$
\mathbf{A}_{\omega\omega} = \begin{pmatrix}
1- \kappa \alpha & 0 & 0\\
0 & 1-\kappa \alpha & 0\\
0 & 0 & 1
\end{pmatrix}
, \quad
\mathbf{A}_{\omega v} = \begin{pmatrix}
0 & -\kappa \frac{\alpha}{r} & 0\\
\kappa \frac{\alpha}{r} & 0 & 0\\
0 & 0 & 0
\end{pmatrix}
$},
\end{align}

where
\begin{equation}
    \alpha = \min \left(\frac{2}{5}, \mu (1 + e_n) \frac{\mid v_z^- \mid}{\| \mathbf{v}_T ^-\|}\right)
\end{equation}
determines if the contact is rolling or sliding, $\kappa = mr^2/I = 3/2$ the inertia ratio for a thin spherical shell of mass $m$, $\mu=0.25$ is the coefficient of dynamic friction, $e_n$ is the coefficient of restitution and
\begin{equation}
 \mathbf{v}_T  = \begin{pmatrix}
v_x\\
v_y\\
0
\end{pmatrix}^-
+\begin{pmatrix}
\omega_x\\
\omega_y\\
\omega_z
\end{pmatrix}^-
\times
\begin{pmatrix}
0\\
0\\
-r
\end{pmatrix}
\end{equation}
is the surface velocity. Superscripts $^-$ and $^+$ indicate pre- and post-contact quantities respectively.

It is an established fact that the restitution coefficient is not constant and that the impact velocity has an effect due to ball buckling \cite{Inaba2017,Haron2012,Rmond2022,Bao2015,Rmond2024} and therefore we update the model to fit our competitive play needs with a coefficient of restitution expressed as a linear function of velocity $e_n = e_0 + e_1\, v^-_z = 0.98 + 0.02 v^-_z$ .

Additionally, a velocity-direction dependent residual correction is applied to account for systematic deviations from the idealized model described above. The correction matrices $\Delta \mathbf{A}_{vv}$, $\Delta \mathbf{A}_{v\omega}$, $\Delta \mathbf{A}_{\omega v}$, $\Delta \mathbf{A}_{\omega\omega}$
are defined in a \emph{local} coordinate frame aligned with the velocity projected on the table plane, then rotated to the global frame and are obtained by Lasso regression~\cite{Tibshirani1996} (L1-regularized least-squares) against experimental table contact data, which induces sparsity in the correction matrices:
% Lasso code: project_ace -- fujiwarac/ace_evaluationWIP:src/utils/ace_evaluation/ace_evaluation/to_maybe_remove/compare_table_contact_datasets.py

% this is the version that has been used so far
%\begin{equation}
%  \Delta \mathbf{A}_{vv}^{\text{loc}} =
%  \begin{pmatrix}
%    -0.01087 & 0 & 0 \\
%    0 & 0 & 0 \\
%    -0.01433 & 0 & 0
%  \end{pmatrix}
%\end{equation}
%\begin{equation}
%  \Delta \mathbf{A}_{v\omega}^{\text{loc}} =
%  \begin{pmatrix}
%    -7.302{\times}10^{-6} &  5.744{\times}10^{-4} &  8.345{\times}10^{-5} \\
%    -5.860{\times}10^{-4} & -3.063{\times}10^{-5} & -1.666{\times}10^{-4} \\
%    -2.485{\times}10^{-5} & -1.322{\times}10^{-4} & -1.428{\times}10^{-5}
%  \end{pmatrix}
%\end{equation}
%\begin{equation}
%  \Delta \mathbf{A}_{\omega v}^{\text{loc}} =
%  \begin{pmatrix}
%    -0.07352 & 0 & 0 \\
%     3.08839 & 0 & 3.03143 \\
%    -0.03828 & 0 & 0.24827
%  \end{pmatrix}
%\end{equation}
%\begin{equation}
%  \Delta \mathbf{A}_{\omega\omega}^{\text{loc}} =
%  \begin{pmatrix}
%     0.01338 &  0.00225 &  0.01107 \\
%    -0.00465 &  0.00189 & -0.00551 \\
%     0.04506 & -0.00348 &  0.05769
%  \end{pmatrix}
%\end{equation}

% this represents the latest update with new parameters
\begin{equation}
  \Delta \mathbf{A}_{vv}^{\text{loc}} =
  \begin{pmatrix}
    0 & 0 & 0 \\
    0 & 0 & 3.378{\times}10^{-3} \\
    -0.02344 & 0 & -0.02717
  \end{pmatrix}
\end{equation}
\begin{equation}
  \Delta \mathbf{A}_{v\omega}^{\text{loc}} =
  \begin{pmatrix}
    0 &  3.330{\times}10^{-4} & 0 \\
    -6.940{\times}10^{-4} & -5.428{\times}10^{-5} & -2.126{\times}10^{-4} \\
     1.768{\times}10^{-5} & -1.148{\times}10^{-5} & -8.228{\times}10^{-6}
  \end{pmatrix}
\end{equation}
\begin{equation}
  \Delta \mathbf{A}_{\omega v}^{\text{loc}} =
  \begin{pmatrix}
    -0.69324 & 0 & -1.02984 \\
     0.51114 & 0 &  0 \\
     0.34033 & 0 &  0
  \end{pmatrix}
\end{equation}
\begin{equation}
  \Delta \mathbf{A}_{\omega\omega}^{\text{loc}} =
  \begin{pmatrix}
     0.06456 &  3.501{\times}10^{-4} &  0.01193 \\
     0.00756 &  0.00211 &  0.00389 \\
     0.00314 &  0.00291 &  0.03411
  \end{pmatrix}
\end{equation}

The corrected matrices are obtained by rotating the local corrections into
the global frame and subtracting:
\begin{eqnarray}
  \tilde{\mathbf{A}}_{vv} &=& \mathbf{A}_{vv} - \mathbf{R}^\top \Delta \mathbf{A}_{vv}^{\text{loc}}\, \mathbf{R},\\
  \tilde{\mathbf{A}}_{v\omega} &=& \mathbf{A}_{v\omega} - \mathbf{R}^\top \Delta \mathbf{A}_{v\omega}^{\text{loc}}\, \mathbf{R},\\
  \tilde{\mathbf{A}}_{\omega v} &=& \mathbf{A}_{\omega v} - \mathbf{R}^\top \Delta \mathbf{A}_{\omega v}^{\text{loc}}\, \mathbf{R},\\
  \tilde{\mathbf{A}}_{\omega \omega} &=& {\mathbf{A}}_{\omega \omega} - \mathbf{R}^\top \Delta {\mathbf{A}}_{\omega \omega}^{\text{loc}}\, \mathbf{R},
\end{eqnarray}

where
\begin{equation}
  \mathbf{R} = \begin{pmatrix}
    v_x / v_{xy} & v_y / v_{xy} & 0 \\
    -v_y / v_{xy} & v_x / v_{xy} & 0 \\
    0 & 0 & 1
  \end{pmatrix}
  \label{eq:table_rotation}
\end{equation}
with $v_{xy} = \sqrt{v_x^2 + v_y^2}$.

The final post-contact state is:
\begin{equation}
    \mathbf{v}^+ = \tilde{\mathbf{A}}_{vv}\,\mathbf{v}^- + \tilde{\mathbf{A}}_{v\omega}\,\boldsymbol{\omega}^-,
    \qquad
    \boldsymbol{\omega}^+ = \tilde{\mathbf{A}}_{\omega v}\,\mathbf{v}^-
                         + \tilde{\mathbf{A}}_{\omega\omega}\,\boldsymbol{\omega}^-
  \label{eq:table_corrected}
\end{equation}

%\paragraph{Remark.}
%For a tilted table surface, the same formulation applies in the table's
%local frame: the pre-contact state is first rotated into the table frame
%via the table orientation quaternion $q_{\text{table}}$, the contact model
%is applied, and the result is rotated back to the world frame.

\subsection{Racket Contact Model}

% We model the ball–racket impact as an instantaneous point collision. 
Let $\mathbf{v}$ and $\boldsymbol{\omega}$ denote the ball linear and angular velocity, respectively, and let $\mathbf{v}_R$ denote the racket velocity. Superscripts $(\cdot)^-$ and $(\cdot)^+$ refer to pre- and post-contact quantities. The racket orientation is described by a rotation matrix $\mathbf{R}_R \in SO(3)$ that transforms vectors from the racket body frame to the world frame.

\paragraph{Base model}
The base model treats the ball–racket impact as an instantaneous point collision. 
We define the relative velocity as $\mathbf{v}_{\mathrm{rel}} = \mathbf{v}^- - \mathbf{v}_R$, where $\mathbf{v}^-$ is the ball velocity before contact and $\mathbf{v}_R$ is the racket velocity at contact. Similarly to the table contact modeling, we start from the model proposed by Nakashima et al. \cite{Nakashima2011}:
\begin{eqnarray}
\mathbf{v}_{\mathrm{rel}}^+
&=&
\mathbf{R}_R
\mathbf{A}_{v v}
\mathbf{R}_R^T
\mathbf{v}_{\mathrm{rel}}^- +
\mathbf{R}_R
\mathbf{A}_{v \omega}
\mathbf{R}_R^T
\boldsymbol{\omega}^-,\\
\boldsymbol{\omega}^+
&=&
\mathbf{R}_R
\mathbf{A}_{\omega v}
\mathbf{R}_R^T
\mathbf{v}_{\mathrm{rel}}^- +
\mathbf{R}_R
\mathbf{A}_{\omega \omega}
\mathbf{R}_R^T
\boldsymbol{\omega}^-,
\end{eqnarray}

% version with racket angular velocity
%\begin{eqnarray}
%\begin{pmatrix}
%\mathbf{v} - \mathbf{v}_R - d\boldsymbol{\omega}_R\\
%\boldsymbol{\omega} - \boldsymbol{\omega}_R
%\end{pmatrix}^+
%=
%\begin{pmatrix}
%    \mathbf{R}_R & \mathbf{0} \\
%    \mathbf{0} & \mathbf{R}_R
%\end{pmatrix}
%\begin{pmatrix}
%\mathbf{A}_{u u} & \mathbf{A}_{u \omega}\\
%\mathbf{A}_{\omega u} & \mathbf{A}_{\omega \omega}
%\end{pmatrix}
%\begin{pmatrix}
%    \mathbf{R}_R & \mathbf{0} \\
%    \mathbf{0} & \mathbf{R}_R
%\end{pmatrix}^T
%\begin{pmatrix}
%\mathbf{v} - \mathbf{v}_R - d\boldsymbol{\omega}_R\\
%\boldsymbol{\omega} - \boldsymbol{\omega}_R
%\end{pmatrix}^-
%\end{eqnarray}
where the coefficient matrices have the same form as in Eq. \ref{eq:tcm_v}  and \ref{eq:tcm_w} , but with $\alpha=\frac{k_p}{m}$ and $k_p$ a coefficient relating tangential velocity to tangential impulse.

\paragraph{Updated base model}
We extend the previous model with three changes. First, the normal restitution coefficient is made velocity-dependent: $e_r = e_0 + e_1\,\|\mathbf{v}_n^-\|$, analogous to the table contact model, where $\|\mathbf{v}_n^-\|$ is component of the velocity normal to the impact surface. Second, $\alpha$ is changed to represent the effective grip fraction $\alpha = (1+e_t)/(1+\kappa)$, where  $e_t = e_{t,0} + e_{t,1}\,\|\mathbf{v}_T\|$ is a tangential coefficient of restitution that depends on the velocity of the surface at contact. Third, we incorporate a normal spin damping coefficient $e_s$ by updating the $(3,3)$ entry of $\mathbf{A}_{\omega\omega}$:
\begin{equation}
\mathbf{A}_{\omega\omega} = \begin{pmatrix}
1-\kappa \alpha & 0 & 0\\
0 & 1-\kappa \alpha & 0\\
0 & 0 & e_s
\end{pmatrix}.
\end{equation}

% with $\kappa = \dfrac{mr^2}{I} = \dfrac{3}{2}$ and 
% \begin{equation}
%     \mathbf{R}_R = \begin{pmatrix}
% \cos\beta & \sin\beta\sin\alpha & \sin\beta\cos\alpha\\
% 0 & \cos\alpha & -\sin\alpha\\
% -\sin\beta & \cos\beta\sin\alpha & \cos\beta\cos\alpha
% \end{pmatrix}.
% \end{equation}
% The models are updated with a linear restitution coefficient $ e_r = e_0 + e_1\,|v_n^-|$ analogous to the one used in the table contact model, a tangential coefficient of restitution $e_t=e_{t,0} + e_{t,1}\,\|\mathbf{u}_T\|$ and normal spin damping $e_s$.
%We also consider effects of racket angular velocity, which can be considerable.
%contact refinement

% In the updated model, spin damping $e_s$ is incorporated as:
% \begin{eqnarray}
% \mathbf{A}_{\omega\omega} = \begin{pmatrix}
% 1- \kappa k_p & 0 & 0\\
% 0 & 1-\kappa k_p & 0\\
% 0 & 0 & e_s
% \end{pmatrix}
% \end{eqnarray}
% and $k_p$ becomes $k_p=\frac{1 + e_t}{1 + \kappa}$, which represents the effective grip fraction.
The five free parameters are fit from experimental racket contact data (and are therefore tailored to the specific combination of racket rubber, blade, and ball used in our experiments) and identified as shown in Table \ref{tab:rcm_params}.

\begin{table}[t]
\caption{Fitted parameters for the racket contact model.}
\label{tab:rcm_params}
\centering
\begin{tabular}{c l l}
\hline
Symbol & Value \\
\hline
$e_0$       & $0.878$ \\
$e_1$       & $-0.020$ \\
$e_{t,0}$   & $0.819$ \\
$e_{t,1}$   & $-0.010$ \\
$e_s$& $0.805$\\
\hline
\end{tabular}
\end{table}

%Note that $k_p \leq 1$ is not explicitly enforced; the clipping of $e_t$ to $[0,\, 1.5]$ ensures $k_p \leq 1$ when $\kappa = 3/2$.

%\paragraph{Remark.}
%When the contact offset from the racket centre is significant, the effective racket velocity at the contact point includes the angular contribution:
%$\mathbf{v}_{\text{rac}} \leftarrow \mathbf{v}_{\text{rac}} + \boldsymbol{\omega}_{\text{rac}} \times (R_{\text{rac}}\,\mathbf{p}_{\text{offset}})$,
%where $\mathbf{p}_{\text{offset}}$ is the contact position in the racket body frame.

\paragraph{Residual neural network}
To further improve prediction accuracy and compensate for non-linear effects that are difficult to model analytically, we train a residual neural network in the racket body frame. These effects arise from the complex ball–racket interaction and the multilayer racket structure, making a neural network an efficient data-driven approach for capturing dynamics not represented by the base model.

\begin{figure}[t]
    \centering
    \includegraphics[width=\columnwidth]{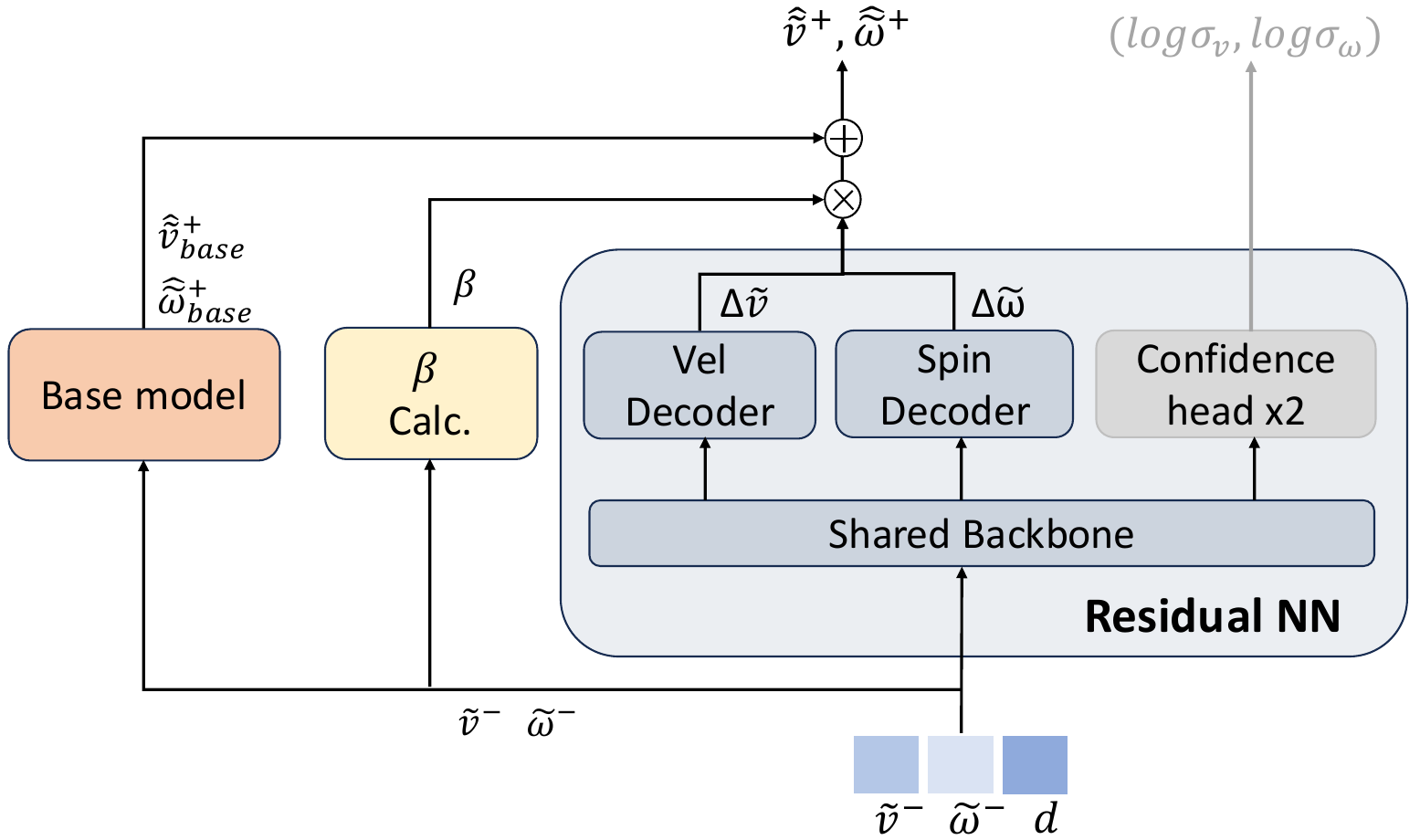}
    \caption{Architecture of the residual neural network. The concatenated racket-frame input $\mathbf{x} = (\tilde{\mathbf{v}}^{-},\, \tilde{\boldsymbol{\omega}}^{-},\, \mathbf{d})$ is passed through a shared backbone, which feeds two decoder heads producing the velocity and spin residuals $(\Delta\tilde{\mathbf{v}},\, \Delta\tilde{\boldsymbol{\omega}})$ and two confidence heads producing per-component log-standard-deviations $(\log\boldsymbol{\sigma}_v,\, \log\boldsymbol{\sigma}_\omega)$. The residuals are combined with the base model prediction via the distance-based attenuation $\beta(\mathbf{x})$ to yield the final post-contact state $(\hat{\tilde{\mathbf{v}}}^{+},\, \hat{\tilde{\boldsymbol{\omega}}}^{+})$.}
    \label{fig:residual_nn_arch}
\end{figure}

Unlike the base model, we explicitly account for the racket angular velocity $\boldsymbol{\omega}_R$ when forming the network input and target. Let $\mathbf{r} = \mathbf{p}_{\mathrm{ball}} - \mathbf{p}_{\mathrm{racket}}$ be the contact offset in the world frame. The pre-contact state is transformed into the racket frame as
\begin{align}
    \tilde{\mathbf{v}}^{-} &= \mathbf{R}_R^{T}\bigl(\mathbf{v}^{-} - \mathbf{v}_R - \boldsymbol{\omega}_R \times \mathbf{r}\bigr), \\
    \tilde{\boldsymbol{\omega}}^{-} &= \mathbf{R}_R^{T}\bigl(\boldsymbol{\omega}^{-} - \boldsymbol{\omega}_R\bigr),
\end{align}
and the network input is the concatenation
\begin{equation}
    \mathbf{x} = \bigl(\tilde{\mathbf{v}}^{-},\; \tilde{\boldsymbol{\omega}}^{-},\; \mathbf{d}\bigr) \in \mathbb{R}^{8},
\end{equation}
where $\mathbf{d} = (d_y,\, d_z) \in \mathbb{R}^{2}$ is the tangential contact position on the racket surface in the racket frame, included to capture property variations across the racket surface such as the coefficient of restitution. The concatenated input passes through a shared trunk of fully-connected layers with batch normalization, ReLU activations, and dropout (Fig.~\ref{fig:residual_nn_arch}). Two parallel decoder heads produce the velocity and spin residuals $(\Delta\tilde{\mathbf{v}},\,\Delta\tilde{\boldsymbol{\omega}}) \in \mathbb{R}^6$ and two additional confidence heads predict per-component log-standard-deviations $(\log\boldsymbol{\sigma}_v,\,\log\boldsymbol{\sigma}_\omega)$, providing heteroscedastic uncertainty estimates.

Letting $\hat{\mathbf{y}}_{\mathrm{base}}(\mathbf{x}) = (\hat{\tilde{\mathbf{v}}}^{+}_{\mathrm{base}},\, \hat{\tilde{\boldsymbol{\omega}}}^{+}_{\mathrm{base}}) \in \mathbb{R}^{6}$ be the updated base model prediction of the post-contact state in the racket frame, the residuals are added to obtain the corrected racket-frame prediction:
\begin{align}
    \hat{\tilde{\mathbf{v}}}^{+} &= \hat{\tilde{\mathbf{v}}}^{+}_{\mathrm{base}} + \beta(\tilde{\mathbf{v}}^{-},\; \tilde{\boldsymbol{\omega}}^{-})\,\Delta\tilde{\mathbf{v}}, \\
    \hat{\tilde{\boldsymbol{\omega}}}^{+} &= \hat{\tilde{\boldsymbol{\omega}}}^{+}_{\mathrm{base}} + \beta(\tilde{\mathbf{v}}^{-},\; \tilde{\boldsymbol{\omega}}^{-})\,\Delta\tilde{\boldsymbol{\omega}},
\end{align}
where $\beta(\mathbf{x}) \in [0,\,1]$ is a distance-based attenuation factor, computed as an exponential decay of the distance between $\mathbf{x}$ and the nearest cluster center of the training data in normalized input space, that smoothly suppresses the learned correction for out-of-distribution inputs, ensuring the model gracefully reverts to the analytical baseline as a safety fallback. The post-contact world-frame state is then recovered by inverting the racket-frame transformation:
\begin{align}
    \mathbf{v}^{+} &= \mathbf{R}_R\,\hat{\tilde{\mathbf{v}}}^{+} + \mathbf{v}_R + \boldsymbol{\omega}_R \times \mathbf{r}, \\
    \boldsymbol{\omega}^{+} &= \mathbf{R}_R\,\hat{\tilde{\boldsymbol{\omega}}}^{+} + \boldsymbol{\omega}_R.
\end{align}

The model is trained end-to-end with the Adam optimizer with decoupled weight decay by minimizing the Gaussian negative log-likelihood:
\begin{equation}
    \mathcal{L} = \frac{1}{N}\sum_{i=1}^{N}\sum_{j=1}^{6}\left[\log\sigma_j(\mathbf{x}_i) + \frac{\bigl(y_{ij}-\hat{y}_{ij}\bigr)^2}{2\,\sigma_j^2(\mathbf{x}_i)}\right],
\end{equation}
which simultaneously penalizes prediction error and overconfident uncertainty estimates, while providing robustness to noisy observations through uncertainty-aware weighting of prediction errors. 

\section{Results}
\label{Results}
We evaluate each proposed model individually against the baselines of Nakashima et al. \cite{Nakashima2011} and D\"urr et al. \cite{Durr2026} before assessing their combined effect on landing position error over full simulated rally segments.

\subsection{Aerodynamics model}
\begin{figure}
    \centering
    \includegraphics[width=\linewidth]{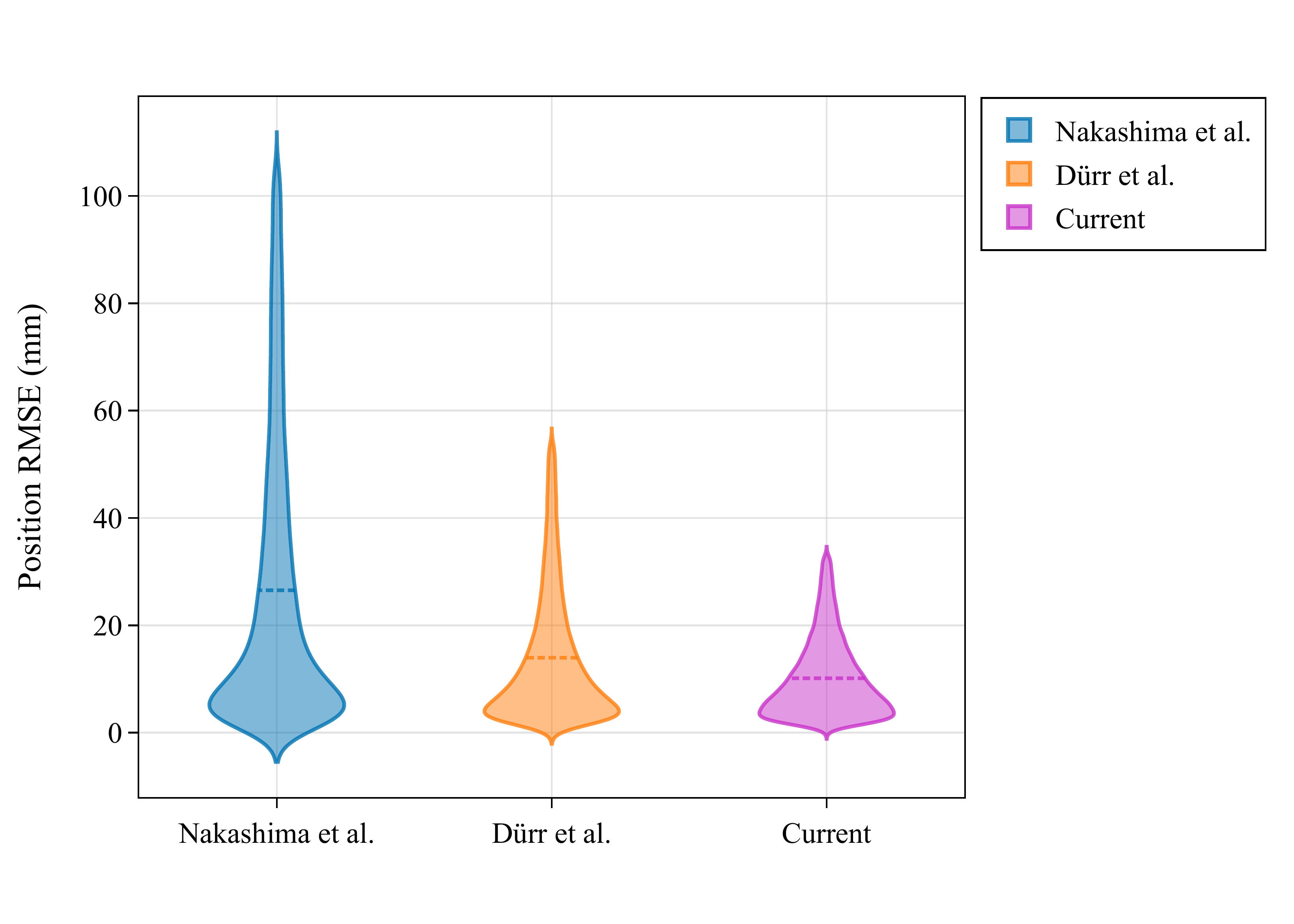}
    \caption{Violin plots of RMSE error for flight trajectories: the blue violin shows the reference model from Nakashima et al. \cite{Nakashima2011}, the orange violin shows the results obtained with the model used in D\"urr et al. \cite{Durr2026} and the magenta violin shows the proposed approach.}
    \label{fig:aero_rmse}
\end{figure}

Fig.~\ref{fig:aero_rmse} shows violin plots of the 3D ball position RMSE — computed over individual flight segments against ball position observations — for the three models compared. The proposed approach (magenta) reduces median RMSE from 15.8 mm to 8.1 mm compared to Nakashima et al. \cite{Nakashima2011} (blue), a 49\% improvement, and from 9.7 mm to 8.1 mm compared to D\"urr et al. \cite{Durr2026} (orange), a 16\% improvement. Improvements at the 75th percentile are larger — 65\% and 28\% respectively — indicating that the proposed model is particularly effective at reducing large errors on difficult trajectories.

\subsection{Table contact model}
\begin{figure}
    \centering
    \includegraphics[width=\linewidth]{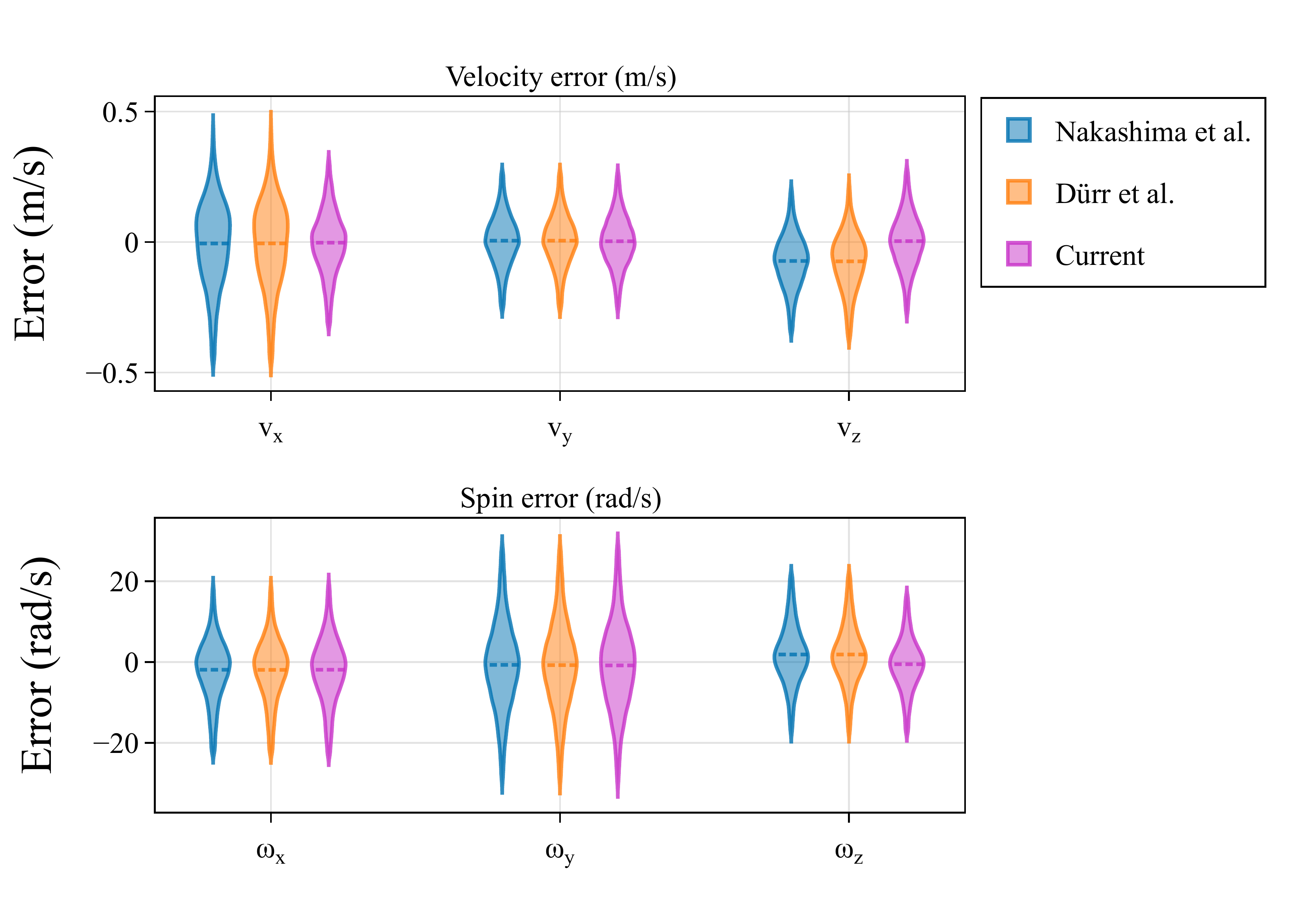}
    \caption{Violin plots for the table contact model per-component errors compared to observations, comparing Nakashima et al. \cite{Nakashima2011} (blue), D\"urr et al. \cite{Durr2026} (orange) and the proposed  approach (magenta).}
    \label{fig:tcm}
\end{figure}

% is there a point in showing the ITTF version?
%Fig. \ref{fig:tcm} shows the per-component improvements brought by the changes proposed in this work compared to previous results from \cite{Nakashima2011} in two variations: one that uses a coefficient of restitution estimated from ITTF requirements on the table and one based on the parameter proposed in the original publication. The error reductions are generally not significant outside of the coefficient of restitution where the results bias is removed.
Fig. \ref{fig:tcm} shows the per-component errors for the table contact model. The largest improvements over both baselines are in $v_x$ (around $30\%$ reduction in median error) and $v_z$ (around $20\%$), where the residual correction effectively removes systematic biases. The $w_z$ component shows a consistent improvement of around $14\%$ across all percentiles, suggesting that the residual model captures a damping effect of the spin.

\subsection{Racket contact model}
\begin{figure}
    \centering
    \includegraphics[width=\linewidth]{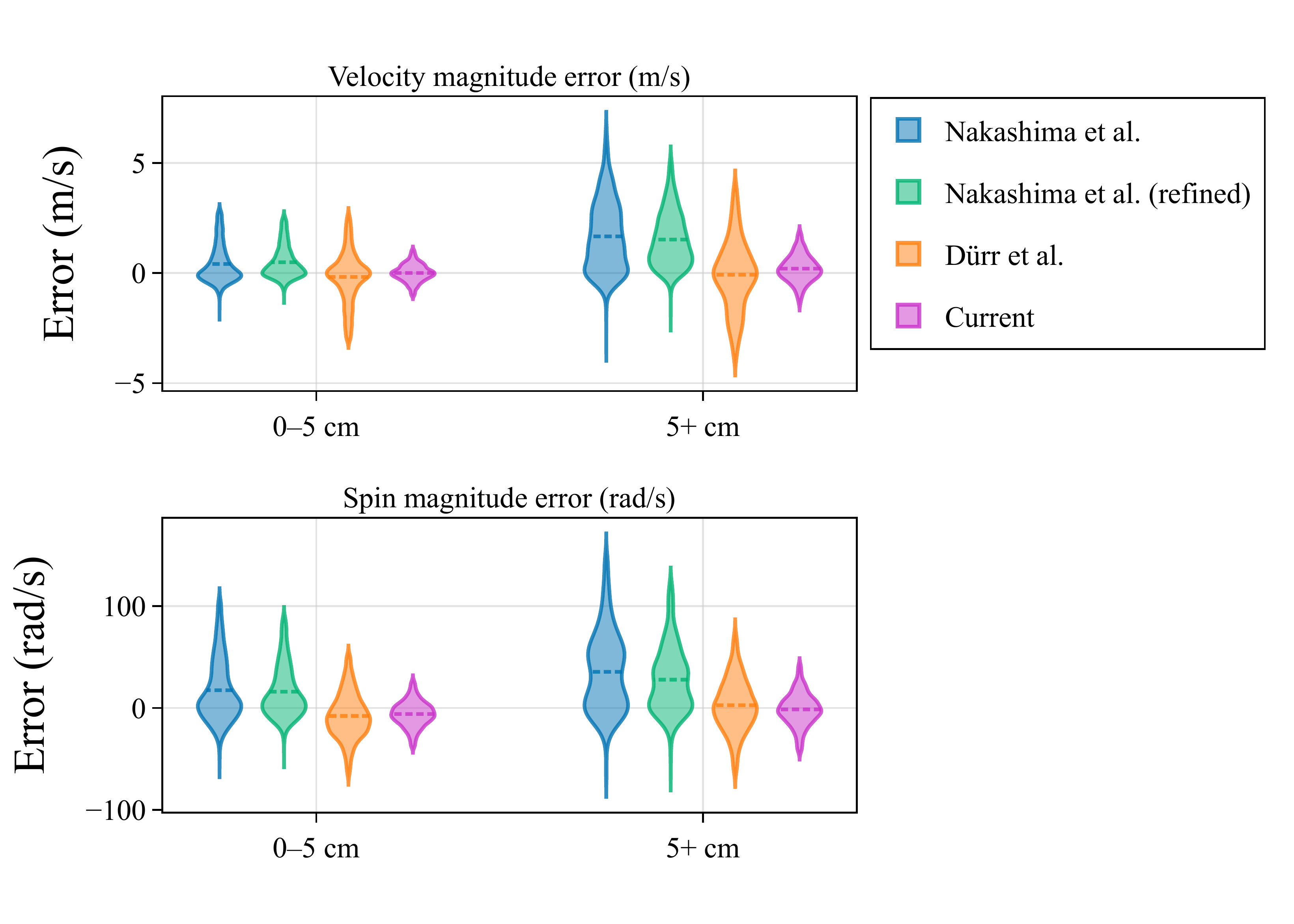}
    \caption{Violin plots for racket contact model errors w.r.t. observations, comparing Nakashima et al. \cite{Nakashima2011} (blue), Nakashima et al. \cite{Nakashima2011} with racket angular velocity (``refined'', green), D\"urr et al. \cite{Durr2026} (orange), and the proposed approach (magenta), grouped by contact distance from the racket center.}
    \label{fig:rcm}
\end{figure}

%Fig. \ref{fig:rcm} shows the performance of the racket contact model from \cite{Nakashima2011} with simple observations as input (blue) and with "refined" quantities, where contact quantities are estimated at sub observation frequency and account for racket angular velocity (yellow) compared to the updated version proposed in this work (magenta).
%The plot shows magnitude errors for velocity and spin, grouped by estimated contact distance from the center of the racket.
Fig.~\ref{fig:rcm} shows the performance of the racket contact model. 
Although the original Nakashima et al.~\cite{Nakashima2011} model does not explicitly account for racket angular velocity, doing so amounts to a coordinate transformation rather than a modeling contribution. We therefore include a ``refined'' variant of Nakashima et al.~\cite{Nakashima2011} that incorporates racket angular velocity (green), isolating its effect from the modeling improvements introduced in this work.
D\"urr et al.~\cite{Durr2026} similarly did not account for racket angular velocity, whereas the proposed model (magenta) does.

The refined Nakashima et al.~\cite{Nakashima2011} baseline shows only modest improvements over the original --- around $1$--$18\%$ depending on the component --- confirming that the coordinate transformation alone is insufficient. Nakashima et al.~\cite{Nakashima2011} exhibits a strong positive bias in both velocity and spin, whereas the models developed in D\"urr et al.~\cite{Durr2026} are more symmetric but with comparable spread; we note that the models from D\"urr et al.~~\cite{Durr2026} were fit on a different dataset and may reflect differences in hardware configuration and player skills between the two works. The proposed model reduces and centers errors across all components, achieving median velocity and spin magnitude reductions of around $43$--$62\%$ and $45$--$49\%$ respectively over both baselines.
In the $0-5$ cm band models should be the most accurate but racket angular velocity can play a role.
In the $>5$ cm band the embossed logo on the racket rubber sides close to the handle (between $5$ and $8$ cm) and racket edges as well as anisotropic properties of the rubber due to asymmetric tension ($>8$ cm) can play an unpredictable effect and thus models are generally expected to perform worse. 
In general the results show that the proposed racket contact model has a much smaller error and removes most of the biases, whereas the reference model has a tendency to over-estimate both velocity and spin.

\subsection{Sim-to-Real}
\begin{figure}
    \centering
    \subfloat[Proposed model.]{\includegraphics[width=0.48\linewidth]{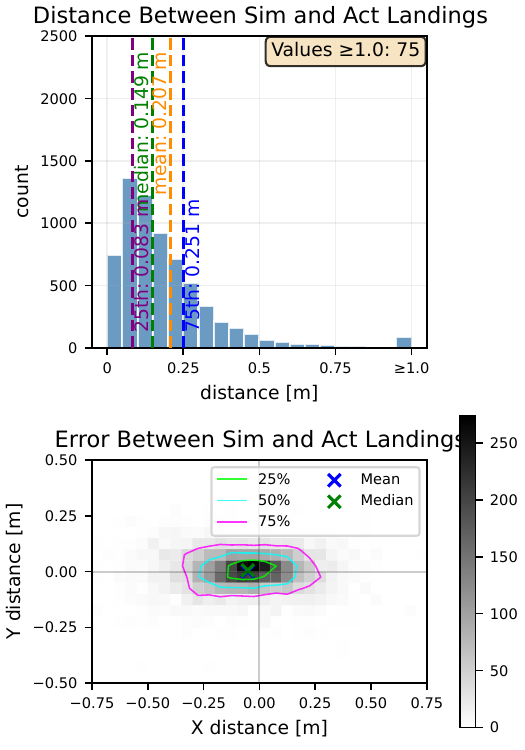}\label{fig:landing_latest}}
    \hfill
    \subfloat[Nakashima et al. \cite{Nakashima2011}.]{\includegraphics[width=0.48\linewidth]{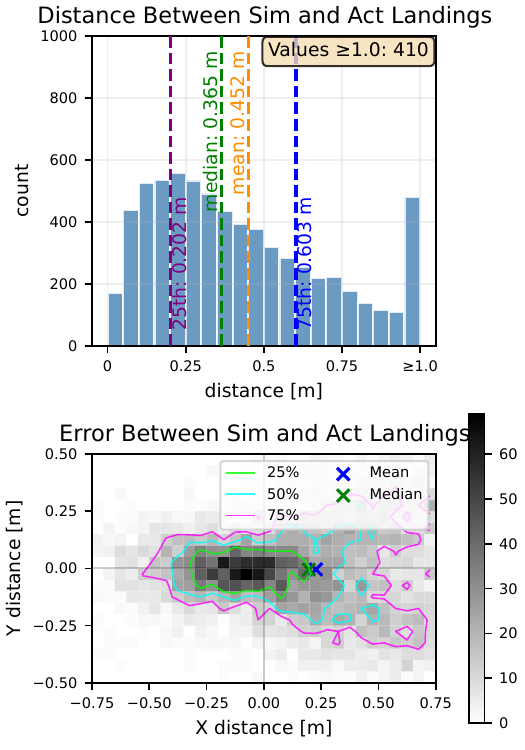}\label{fig:landing_ref}}
    \caption{Landing location accuracy of racket contact model and aerodynamics model. The top plots show the histogram of the error distribution of the predicted landing point compared to observations. The bottom plots show the distributions in XY.}%{\color{red}TODO - the bottom plots might be sufficient to save space?}}
    \label{fig:landing}
\end{figure}

Fig. \ref{fig:landing} shows error in landing position when simulating from the last observation before the racket contact until the ball reaches the table plane (even if it does not hit the table), thereby, evaluating the combined effect of the racket contact and aerodynamics models.
We observe that the 75th percentile error is within $25$ cm of the target, which, while large, is a considerable improvement over the $60$ cm given by the reference model in our setting. The proposed model reduces the median landing error from $0.37$ m to $0.15$ m compared to Nakashima et al. \cite{Nakashima2011} (a $59\%$ reduction), representing the difference between just returning the ball and being able to aim at regions of the table.
The aerodynamics alone accounts for a very small part of this error (compare with Fig. \ref{fig:aero_rmse}, where the RMSE is generally below $2$ cm) and therefore the major cause of the sim-to-real gap in this case is the racket contact model.
An important part of out-of-target shots is caused by the current lack of control on the agent side on where the ball-racket contact happens: as shown in Fig. \ref{fig:rcm}, when the ball hits far from the center of the racket, the performance deteriorates significantly due to contacts with racket edges and embossed features on the rubber as well as the anisotropic characteristics of the rubber.
%{\color{red}TODO - About 10\% of observed trajectories hit the net. Trajectories that hit the net either in sim or obs are excluded from these statistics but are visible in the cumulative probability (red in the plots)}
% note that RCM and aerodynamics are more important: during ball arrival, the policy can still correct for errors, but after hitting the ball there's no more corrections possible, and thus it has to be as accurate as possible. Once the ball lands on the table, better table contact accuracy can help tactical choices though

\section{Conclusion}
\label{conclusion}
In this work we propose physics models for the simulation of table tennis in a virtual environment that is used to train RL policies that run in the real world and play competitively.

We introduced improved drag and Magnus coefficients, restitution models, and residual corrections based on a large dataset of competitive play, and showed that each reduces prediction error over standard baselines — closing sim-to-real gaps that, at professional level, would otherwise be exploited by skilled opponents.

Several directions remain open for future work. The racket contact model is the most critical component for further improvement, as racket contact represents the dominant source of sim-to-real error. One promising direction is to improve uncertainty estimation across the racket surface — for instance, learning that contacts near the embossed logo or racket edges carry higher uncertainty, which could enable more principled handling of off-center hits. A second direction is to extend the approach to different rubber types, either by generalizing the current models or by recalibrating them for different equipment configurations. Finally, the racket state estimate used in this work is obtained from motor encoder readings via rigid-body kinematics; manufacturing tolerances, calibration errors, and structural deflections can cause deviations on the order of $8$ mm, $0.2$ m/s, and $0.7$° in position, velocity, and orientation respectively, and incorporating a direct racket state measurement into the simulator represents an important next step.

%In this work we obtained the racket state (position, orientation, linear velocity, angular velocity) from rigid-body forward kinematics applied to the measured motor encoder positions. Due to manufacturing or assembly tolerances, calibration issues, and deflections or vibrations in the robot structure, the true racket state may deviate significantly from this estimate. Using an OptiTrack system we measured that the deviations in racket position, velocity, and orientation can be on the order of 8 mm, 0.2 m/s, and $0.7^{\circ}$, respectively.

%These deviations can contribute significantly to the sim-to-real error of the racket contact model and thus modeling the phenomena and incorporating it into the simulator is an important next step in reducing the sim-to-real gap.

%% I think this is already some partial results that are not useful in this work
%A dataset containing of OptiTrack racket state data was used to train a Convolutional Neural Network that could estimate the velocity and orientation deviation of the racket state due to vibration and deflection given a finite history of joint position reference commands sent to the robot.

% Additional work could also be done to address some of the sources of uncertainty mentioned in \ref{section:uncertainties}, although we believe most should be considered as aleatoric uncertainties to 

%\section*{Acknowledgment}


\bibliographystyle{IEEEtran}
\bibliography{ace_physics}

\end{document}